\documentclass{article}

\usepackage[final]{corl_2018}

\usepackage{graphicx}
\usepackage{booktabs}
\usepackage{wrapfig}
\usepackage{multicol}
\usepackage{subcaption}
\usepackage{xspace}
\usepackage{times}
\usepackage{amsfonts}
\usepackage{latexsym}
\usepackage{url}
\usepackage{wrapfig}
\usepackage{latexsym}
\usepackage{multirow}
\usepackage{mathtools}
\usepackage{calc}
\usepackage{amsmath, amsthm, amssymb, amsfonts}
\usepackage{url}
\usepackage{framed}
\usepackage{textcomp}
\usepackage{mathtools}
\newbox{\bigpicturebox}
\usepackage[export]{adjustbox}
\usepackage{natbib}
\usepackage{url}
\usepackage{soul}
\newcommand{\D}{MIME\xspace} 

\newcommand{\xxnote}[3]{}
\ifx\hidenotes\undefined
  \renewcommand{\xxnote}[3]{\color{#2}{#1: #3}}
\fi

\title{Multiple Interactions Made Easy (MIME) :\\
Large Scale Demonstrations Data for Imitation}

\author{
  Pratyusha Sharma\thanks{Equal contribution. Correspondence: \{{pratyuss, lerrelp, abhinavg}\}{@cs.cmu.edu}, {lekhawm@gmail.com} } \qquad  Lekha Mohan\footnotemark[1] \qquad  Lerrel Pinto \qquad  Abhinav Gupta\\ \\ 
  The Robotics Institute\\
  Carnegie Mellon University
}
\graphicspath {{./figures/}}
\begin{document}
\maketitle


\begin{abstract}
In recent years, we have seen an emergence of data-driven approaches in robotics. However, most existing efforts and datasets are either in simulation or focus on a single task in isolation such as grasping, pushing or poking. In order to make progress and capture the space of manipulation, we would need to collect a large-scale dataset of diverse tasks such as pouring, opening bottles, stacking objects etc. But how does one collect such a dataset? In this paper, we present the largest available robotic-demonstration dataset (MIME) that contains 8260 human-robot demonstrations over 20 different robotic tasks\footnote{Website: https://sites.google.com/view/mimedataset}. These tasks range from the simple task of pushing objects to the difficult task of stacking household objects. Our dataset consists of videos of human demonstrations and kinesthetic trajectories of robot demonstrations. We also propose to use this dataset for the task of mapping 3rd person video features to robot trajectories. Furthermore, we present two different approaches using this dataset and evaluate the predicted robot trajectories against ground-truth trajectories. We hope our dataset inspires research in multiple areas including visual imitation, trajectory prediction and multi-task robotic learning.
\end{abstract}


\section{Introduction}
One of the biggest success stories in recent AI research is the emergence of data-driven approaches. And as we can expect a key ingredient in these data-driven approaches is {\bf DATA} itself. In computer vision, for example, the emergence of ImageNet~\cite{deng2009imagenet} was a key moment for data driven methods like ConvNets. In recent years, data-driven approaches have started to gain momentum in the field of robotics as well. For example, \citet{pinto2016supersizing}, displayed how data collection can be scaled to 50K examples for tasks such as grasping and how deep learning approaches improve with increasing amounts of data. Since then, data-driven algorithms have been scaled up in terms of number of datapoints~\cite{handeyecoord} and shown to be useful for other task such as poking~\cite{pulkitpoking}.

Most of the existing robotics datasets focus on a single task in isolation such as grasping \citet{pinto2016supersizing,handeyecoord}, pushing~\cite{finn2016unsupervised,pintomultitask2017}, poking~\cite{pulkitpoking} or knot-tying~\cite{pulkitrope2017}. This is not surprising; even the early computer vision datasets initially focused on single tasks such as faces~\cite{Turk:1991:ER:1326887.1326894} and cars~\cite{agarwal2002learning}. But the real success of computer vision came from building datasets than span across hundreds and thousands of categories. What the diversity of data allowed was to learn a generic visual representation that could then be transferred for variety of tasks. Inspired from this observation, we argue that it is critical for data-driven manipulation algorithms to be given diverse data of manipulation tasks with hundreds of different objects. But how do you design and collect a large-scale dataset of diverse robotic manipulations?

Unlike existing large-scale datasets which focus on simple tasks, self-supervision via random exploration is unlikely to succeed for complex manipulation tasks like stacking objects. Therefore, learning complex-manipulation requires supervised learning (demonstrations) rather than self-supervision. But to truly scale the abilities of our agents, we will need to scale the amount of demonstration data available. In this work, we take this first step towards creating large-scale demonstration data: specifically, we collect the largest available robotic demonstration dataset (\D) that contains \textbf{8260} \textit{human-robot} demonstrations with over \textbf{20} different robotic tasks. These tasks range from the simple task of pushing objects to the difficult task of stacking household objects. Collecting a dataset of this scale involves several challenging questions: (a) What type of data do we collect?; (b) How do we collect this data?; and (c) How do we ensure that the collected data is meaningful?

One of the key design decisions in creating \D is the mode of getting expert demonstrations. Although several forms of Learning from Demonstration (LfD) data collection strategies are present in literature, we select two ways of capturing demonstrations: (a) Kinesthetic Demonstrations~\cite{argall2009survey}: a kinesthetic method for collecting demonstrations since it allows a human demonstrator to both express their desired motion as well as stay in the constrains of the robot; (b) Visual Demonstration~\cite{simonyan2014two}: Instead of only recording kinesthetic data, we also record a visual demonstration of how humans perform the same tasks. Hence for every kinesthetic robot demonstration, we also collect a corresponding video of the same human demonstrator performing the task with their hands. 

The next challenge is the process of physically collecting these diverse demonstrations. For small scale datasets, often a single expert demonstrator collects all the data. However to really scale demonstration data, we need to get demonstrations from multiple human demonstrators. This comes with both advantages and disadvanges. The advantages are the kinesthetic data not being biased by a single demonstrator's peculiarities and the potential ability to parallelize data collection. The disadvantage however is that human demonstrators who haven't worked with the robot before find it hard to give kinesthetic demonstrations. We solve this challenge by carefully training the demonstrators and follow it by crosschecking the data by other human demonstrators.  

While we foresee multiple ways of using this dataset (e.g, learning action representation to initialize RL), in this paper, we introduce and focus on the task of learning a mapping from visual demonstrations to robotic trajectories. Specifically, given the video of human demonstration, the goal is to predict the joint angles to achieve the same goal. However, there might be multiple possible robot trajectories that can lead to same goal state; therefore during test we collect multiple demonstrations for the same task and use the min-distance from the set of test trajectories as the evaluation metric. These experiments further validate the utility of \D.

\begin{figure}
\centering
  \includegraphics[width=5.5in]{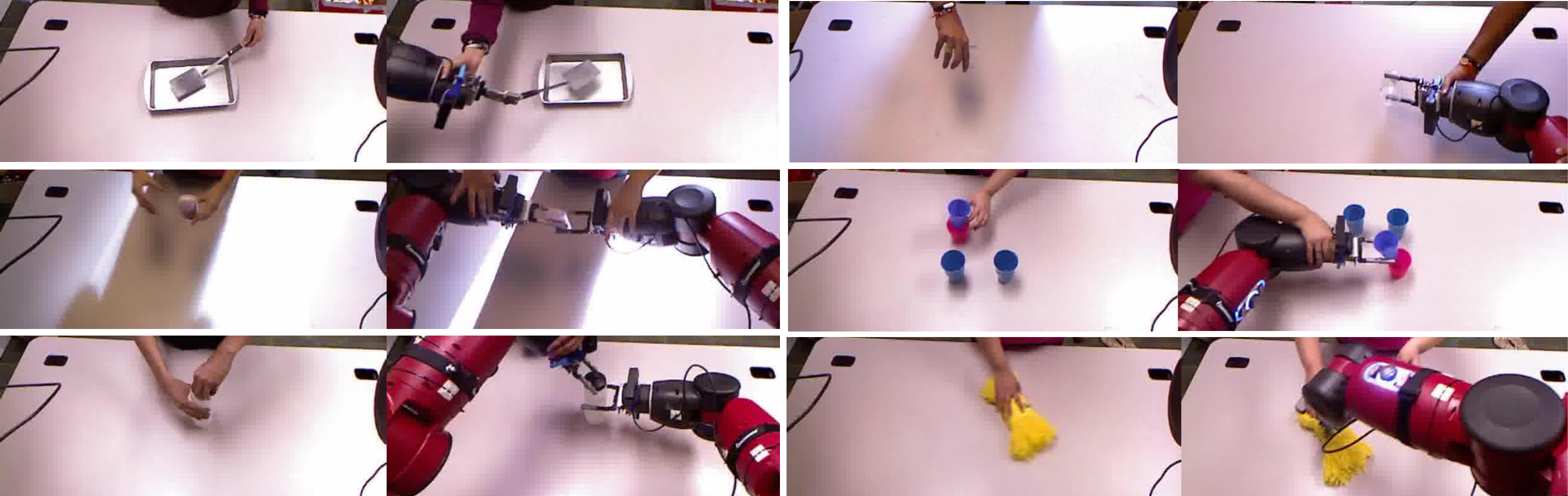}   
\caption{Example datapoints from six diferent task-categories in our MIME Dataset. On the left we show human demonstration and on the right we show how kinesthetic trajectories are being collected.(Clockwise from top-left the tasks are stirring, pouring, stacking, wiping, opening a bottle and passing.) }
\label{fig:human-robot demonstration}
\end{figure}

\section{Related Work}
\subsection{Data Driven Robotics}
Inspired from the successes in large scale computer vision datasets~\cite{deng2009imagenet,Everingham2010}, natural language datasets~\cite{marcus1994penn} and reinforcement learning frameworks~\cite{DBLP:journals/corr/BrockmanCPSSTZ16,todorov2012mujoco}, the last few years has seen a growing interest in large scale robotics~\cite{pinto2016supersizing, handeyecoord}. 
The KITTI driving dataset~\cite{Geiger2013IJRR} collected images from a car that inspired research in several tasks like 3D tracking, visual odometry and optical flow. Similarly the RGBD SLAM dataset~\cite{sturm12iros} has spurred several papers that have improved the state of the art in SLAM. We hope that \D will similarly accelerate research in Learning from Demonstrations.

In the realm of learning from manipulator data, \citet{lenz2015deep} presented one of the first attempts to collect grasping data using expert annotations. \citet{pinto2016supersizing} took the next leap in large scale data collection where the robots collect data themselves without human supervision. This has been followed by attempts to scale even further using multiple robots~\cite{handeyecoord}, multiple tasks~\cite{pinto2016curious}, adversarial learning~\cite{pinto2017supervision} and curriculum learning~\cite{murali2017cassl}. Following grasping, several researchers have looked at the task of pushing objects\cite{pulkitpoking,pintomultitask2017,finn2016unsupervised}. Attempts for large scale data collection has also been pursued in rope manipulation~\cite{pulkitrope2017}, surgical robotics~\cite{gao2014jhu} and opening doors~\cite{gu2017deep}. However all of these tasks look at single and often simple tasks like grasping or pushing objects. In \D, we collect data for around 20 different tasks, which will accelerate robot learning not just for the individual tasks but for imitation learning in general.

Another direction in scaling up data is in robotic simulators. Simulators offer large scale data that can be collected much faster than in the real world. GraspIt!~\cite{miller2004graspit} developed an interface to evaluate robotic grasps in a simulator. DexNet~\cite{mahler2016dex,mahler2017suction} took this further by using cloud computing and synthetic object models to learn grasp models. In driving, several simulators like CARLA~\cite{dosovitskiy2017carla} and AirSim~\cite{shah2018airsim} promise easier self driving research. For indoor navigation simulators like AI2-THOR~\cite{zhu2017target}, SUNCG~\cite{song2017semantic} and Matterport3D~\cite{chang2017matterport3d} have emerged. Simulators can also be interleaved with reinforcement learning for faster learning. Several works~\cite{sadeghi2016cad2rl, pinto2017asymmetric} show promise in this direction. However, transferring policies from simulators to the real world is often challenging due to the reality-gap. This prompts us to create a real-world dataset that would allow for better transfer in the real world.

\subsection{Learning From Demonstrations (LfD)}
Learning from demonstrations encapsulates the field of learning robotic strategies or policies from human or expert demonstrations. An in-depth survey of LfD can be found in \citet{argall2009survey,kober2013reinforcement}. Compared to self-supervized robot learning, LfD methods allows for learning more complex policies. Intuitively, an expert demonstration vastly cuts down the exploration space and can provide strong guidance during policy learning~\cite{kober2013reinforcement}. This has enabled autonomous helicopter aerobatics~\cite{abbeel2010autonomous}, table-tennis playing~\cite{mulling2013learning} and drone flying~\cite{ross2013learning}. However these methods often focus on learning from a handful of expert demonstrations for a single task.

Scaling expert demonstrations has been receiving interest recently with \citet{zhang2017deep} presenting an approach that using tele-operation to collect demonstrations. Here hundreds of demonstrations are collected using a Virtual Reality interface for 15 tasks. However, this data isn't public and used in a task specific manner. For general purpose demonstrations, kinesthetically moving the robot ensures that the robot is in good configuration spaces. This allows for easier whole arm manipulation rather than only end-effector manipulation. Imitation learning using data from multiple tasks has also shown promise~\cite{finn2017one}. We believe that our dataset \D can be used to further research in this area with more diverse objects and larger complexity of tasks.

Learning from demonstrations with multiple tasks is also connected to multi-task learning in the domain of computer vision. Large scale datasets like ImageNet~\cite{deng2009imagenet} allowed for single models to simultaneously classify for multiple categories. These pre-trained classification models~\cite{krizhevsky2012imagenet} can be then used to speed up learning for other visual tasks like detection~\cite{girshick2014rich} and action classification~\cite{simonyan2014two}. However the available large scale visual datasets do not contain physical actions. This makes it hard to transfer learned information to new tasks. Since \D contains rich visual information for 20 different tasks, we believe it will accelerate the progress in multi-task learning from demonstrations.

\section{ The \D Dataset}
We now describe our methodology to collect the MIME demonstration dataset. There are several challenges in this effort: First, what is the right vocabulary of tasks? Second, how do we scale up collection of kinesthetic trajectories? Finally, how do we correct the errors made by humans in demonstration collection?

\subsection{Vocabulary of Tasks}

\begin{figure}
\centering
  \includegraphics[width=5.4in]{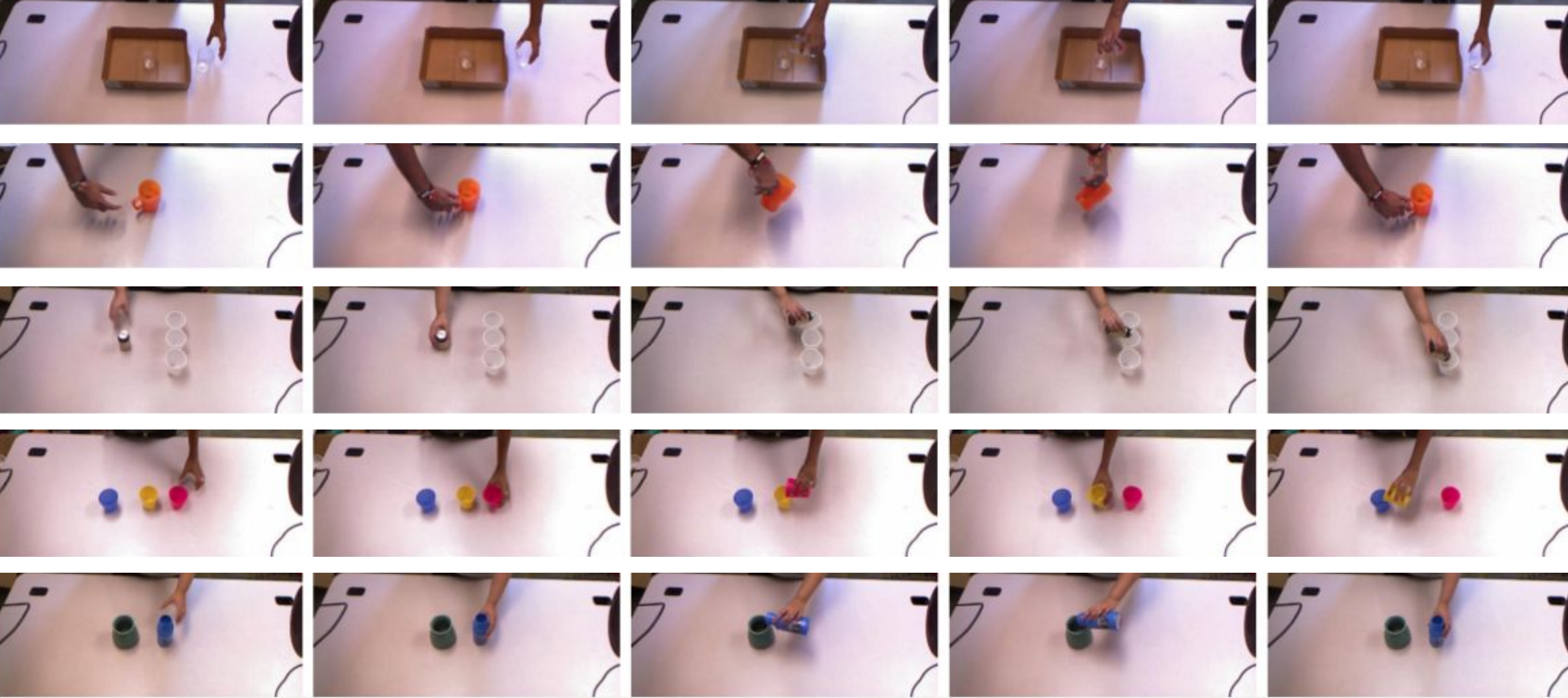}   
\caption{Diversity within a task. For example, in the task of pouring there is diversity in objects (shape, opaque/transparent), arrangement of objects, and how many objects are being poured into.}
\label{fig:diversity}
\end{figure}

The first question that we need to tackle is the vocabulary of tasks for which we collect both the kinesthetic trajectories and the video of human demonstration. The key consideration in selection of tasks is: (a) tasks should be easy enough to be performed by our Baxter robot; (b) should not require haptic feedback for successful performance; (c) diverse enough for us to learn the embedding of tasks. A complete list of the tasks can be seen in Table~\ref{tab:splits}. Finally to increase the diversity of the data, we collect demonstrations over a variety of objects which can be seen in Fig.~\ref{fig:diversity}.

\begin{table}[htb]
\caption{Task-wise data splits}
\begin{tabular}{lllll} \toprule
Task & train & val & test &  \\\midrule
Pour          &  231     & 29    &  30    &  \\ 
Stir         &  406     & 51    &    51  &  \\ 
Pass    & 388      &  48   &  49    &  \\ 
Stack         &   411    &  51   &   52   &  \\ 
Place objects in box & 293       &   37  &  37    &  \\ 
Open bottles  &   378    & 47    &   48   &  \\ 
Push       &  304      & 25     &   26   &  \\ 
Rotate &  335     &  42   &  42    &  \\ 
Wipe            &  233     & 29    &   30   &  \\ 
Press buttons          &  252     & 31    &  32    &  \\
\end{tabular}
\quad
\begin{tabular}{lllll} \toprule
Task & train & val & test &  \\\midrule
Close book          &   253    &  32    &  32    & \\ 
Pick (single hand)        & 524      & 65     &   66   &  \\ 
Pick  (both hands)   &  199     & 25    &   25   &  \\ 
Poke         &  433     & 54    &    55  &  \\ 
Pull (two hands) & 372      &   46  &   47   &  \\ 
Push (two hands)  &   315    &  39   &  40    &  \\ 
Toy car trajectories       &    334   &  42   &   42   &  \\ 
Roll  & 340      &  43   &    43  &  \\ 
Drop objects           &   372    &   46  &  47    &  \\ 
Pull (single hand)            & 328      & 41     &   41   &  \\
\end{tabular}
\label{tab:splits}
\end{table}

\subsection{Robot Setup}
\begin{figure}
\centering
  \includegraphics[width=5.2in]{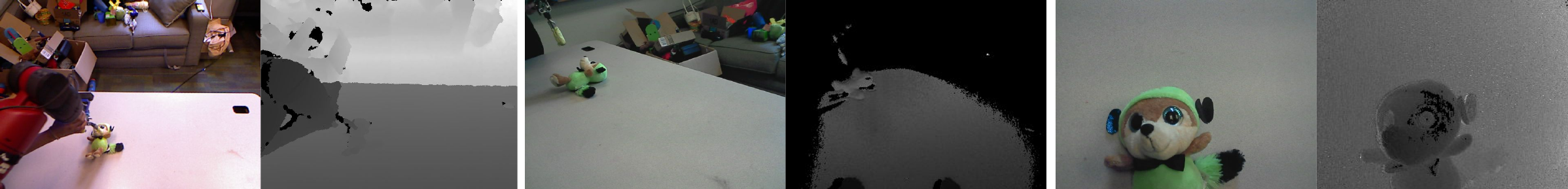}   
\caption{Multiple views of RGBD data from a robot demonstration.}
\label{fig:multi-view}
\end{figure}

To collect the demonstration data, we use a Baxter robot in gravity compensated kinesthetic mode. The Baxter is a dual arm manipulator with 7DoF arms equipped with two-fingered parallel grippers. Furthermore, the robot is equipped with a Kinect mounted on the robot's head, and two SoftKinetic DS325 cameras, each mounted on the robot's wrist. The head camera acts like an external camera observing the task on the table, while the wrist cameras act as eye in robot cameras that move as the arm moves. During every robot demonstration, all the RGBD images as shown in Fig[\ref{fig:multi-view}], the robot joint angles and the gripper positions are synchronized and stored.

\subsection{Data Collection Procedure}

To collect a large scale demonstration dataset, we need mutiple demonstrators to collect data. This brings about a unique challenge: new users often find it difficult to kinesthetically operate the robot. Hence we first train every human demonstrator with lessons to safely handle the robot. Once a demonstrator is comfortable in operating the robot, a specific task is assigned to the participant. 

To facilitate smooth data collection, the participants use specific push buttons on the robot, which in turn displays the instructions on steps to follow. The main features of this setup are to home the robot, record data, visualize successive steps in the demo and display hardware or software errors so the setup can be checked before proceeding. Demonstrations performed by the participants are reviewed by other participants, who visually verify the collected trajectories using a visualizer we made. Using certain keyboard keys the reviewer can accept/reject the demo. If rejected, the trajectories are checked and redone by the demonstrator.

 


For improved robustness and to avoid bias, every participant gives multiple demonstrations for the same task using a variety of objects. After completing demonstrations for a particular task, the participant is assigned a new task. This ensures that each task has multiple demonstrators collecting data for it. The data collection pipeline was iteratively improved based on participant feedback.

Note every data point consists of a human demonstration and a corresponding robot demonstration. We then use a verification stage to check the quality of collected data. Specifically, all the collected data was reviewed by other participants and incorrect demonstrations were removed from the dataset. An interesting observation is that the erroneous trajectories were mostly collected by newer participants. After a few trials, the participants developed the ability to accurately manoeuvre the robot, resulting in less errors and faster trajectories. At the end of the data collection process, we have 8260 demonstrations for 20 discrete tasks (Table \ref{tab:splits}).

\section{Experiments}
We demonstrate the quality of \D by performing a suite of tasks. These tasks will also highlight the importance of the various components of our dataset. These tasks are: Task recognition, and Behaviour Cloning.

To evaluate the algorithms we split the dataset into a train set, validation set, and test set as seen in  Table~\ref{tab:splits}. The distribution of the different task categories is the same over the three splits, with 80\% of the data is in the training set, 10\% in the validation set and 10\% in the test set. However for each of the test-set demonstrations we collect multiple test trajectories. This captures the multi-modal nature of our problem and allows us to compute the error by estimating the distance between predicted trajectories and a set of ground-truth trajectories. 

\subsection{Task Recognition}
We first evaluate on task recognition, where the goal is to classify trajectories based on joint information alone. This is done to demonstrate the discriminative signal in the joint angle data.Since joint trajectories are sequential in nature with trajectory $\tau\equiv(s_0,s_1,..s_{T-1})$, we employ two methods to evaluate on this task, dynamic time warping and a long short term memory architecture (LSTM)~\cite{Hochreiter:1997:LSM:1246443.1246450} as our model. While DTW analytically computes the distance between the time series data it can be slow to use it over large datasets. Hence, we also run a learning based method which captures the knowledge of the data in its weights. 

\noindent \textbf{Dynamic Time Warping(DTW):}  DTW is an algorithm that is often used to look at the proximity of different time series data. We employed DTW between each of the trajectories in the test data with all the trajectories in the training data. We then classified each of the test trajectories into the class of the trajectory of the training data with which it's distance was the least. In this experiment, we again highlight the importance of large-scale data but for the non-learning approach (DTW). Here we see that using just 10\% of data yields an accuracy of 58.9\%, while using the full 100\%, we get a significantly higher accuracy of 79.7\%.

\noindent \textbf{Learning Methods:} We use a single layer LSTM cell followed by a linear layer. At every timestep, the observed state along with the learned hidden state is used to update the prediction of task class. The final prediction, when the last observed state is fed in, is used as the predicted task label. Note that $s_t$ represents the vector of joint angles at time $t$ and $T$ is the length of our trajectory. Our LSTM is unrolled for $T$ steps and fed $s_t$ as input at timestep $t$. The output at every step is a 20 dimensional vector representing the probability of the predicted task. A cross entropy loss is computed at every timestep and cumulated to a final loss. This final loss is minimized using the Adam Optimizer~\cite{kingma2014adam}.





\begin{figure}[htbp]
\begin{center}
\includegraphics[width=4.5in]{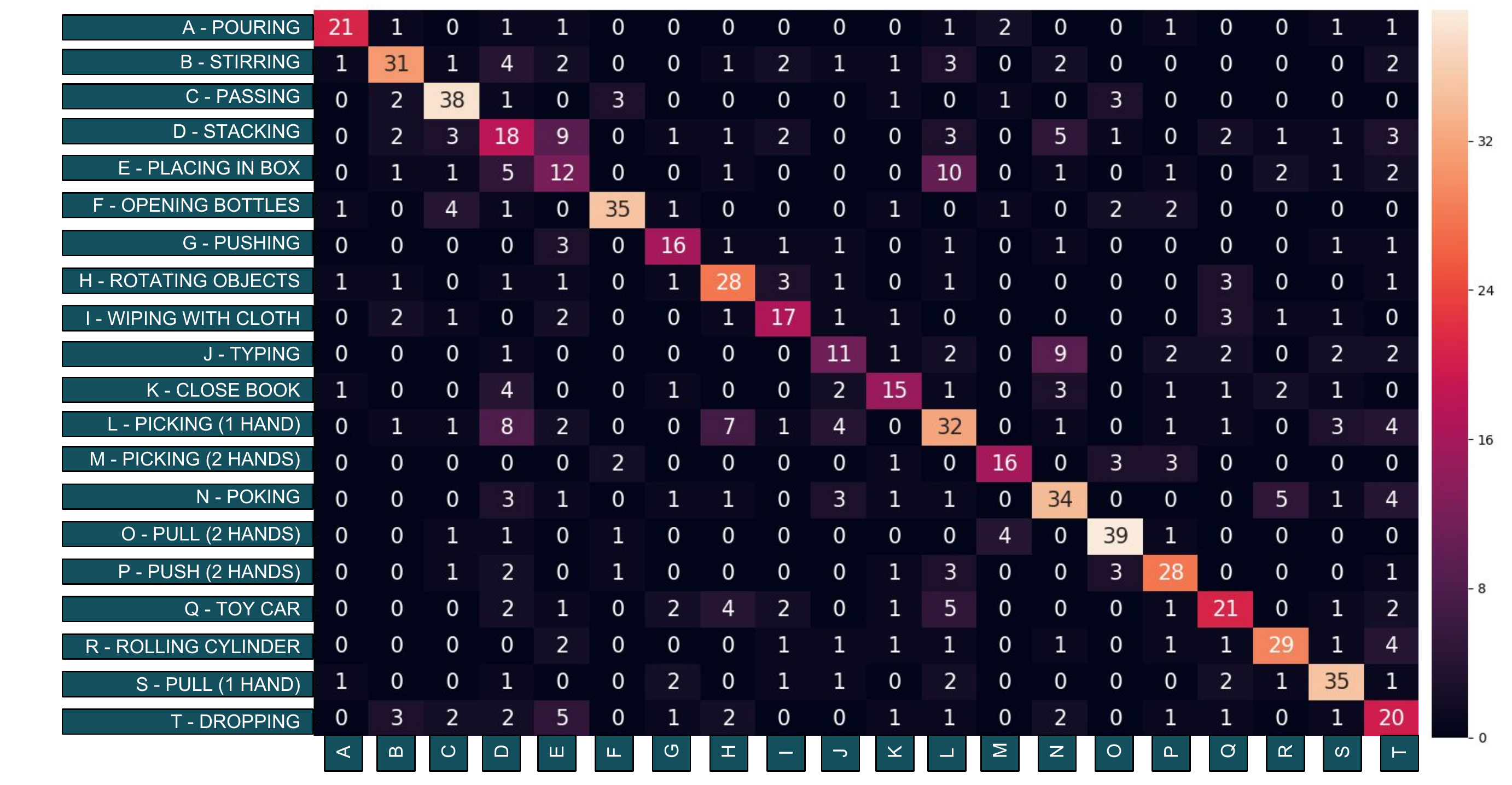}
\caption{Confusion matrix of task classification from joint angles.}
\label{fig:confusionmatrix}
\end{center}
\end{figure}

On the held-out test set, we compare the predicted label at the end feeding in a trajectory with the true label. This LSTM model achieves an accuracy of 61.2\%. Furthermore visualizing the confusion matrix (Fig.~\ref{fig:confusionmatrix}) depicts an interesting trend of task correlation. In most cases, an incorrectly classified datapoint is often from a similar task that has similar or overlapping trajectories. For example picking with a single hand is confused with stacking and placing in a box is confused with picking. 

\subsection{Behavioural Cloning}
One of the challenging tasks we plan to handle using this dataset is behaviour cloning. Specifically, in this task: given the third-person video of a human doing a demonstration and the corresponding robot trajectory for the same task, we want to learn a mapping between visual features to robot trajectories. We will describe the data, a baseline formulation and evaluation metric in the next section.



\noindent \textbf{Data Preparation:} Since the trajectory sequences of different tasks are of different lengths, we sub-sample the robot joint angle trajectories and the human demonstration video frames to a fixed constant length of 50. 

\noindent \textbf{Extracting visual features from demonstration:} Next, we need to extract visual features for the input video. One possibility is to use i3D~\cite{DBLP:journals/corr/FeichtenhoferPW16} or non-local neural network from \citet{Wang_nonlocalCVPR2018}. However, in both these cases, the temporal information is lost in the final features. Therefore, instead for each video we extract 50 sub-clips; each sub-clip is 2-frame clip. We extract features for each sub-clip using a pre-trained model from \citet{Wang_nonlocalCVPR2018}. This leads to a temporally ordered set of 50 sub-clip features.

\noindent \textbf{State Vector:} Apart from using the demonstration features, our imitation policy needs to observe the current object location and state to predict the action plan. To capture the current state, we use a single head-camera image from the robot demonstration instead of the whole video. This image is passed through a pre-trained VGG network~\cite{DBLP:journals/corr/SimonyanZ14a} to obtain image features.


\noindent \textbf{Output Space:} One possibility is to use the current state vector and video features to regress to joint angles directly. However, we note that for each demonstration there are multiple ways to imitate it and hence the multimodal nature of output. Therefore, instead of direct regression, which might regress to the mean of the multimodal trajectories possible, we plan to use a classification-based approach. 

Specifically, first we use the set of joint angles $s_t$ and cluster them using k-means. This allows vector quantization of the output space and hence the model predicts the output cluster center at each time instant. 
But one issue with the above is that if we use one vector-quantization across all possible joint angles, fine-grained manipulation motion tend to be clustered in one group leading to the averaging effect. Therefore, instead, we follow a variable sparsity approach for clustering. Since any trajectory involves first approaching the object and then interacting with it, we cluster the first half of the trajectories ($t\leq25$) sparsely while densely clustering the second half ($t>25$). This helps us better capture the nuances of manipulating the object by finer clustering of the prediction space near the objects. In total we obtain 600 clusters with 150 in the sparse region and 450 in the dense region.



\begin{figure}[htbp]
\begin{center}
\includegraphics[width=4.3in]{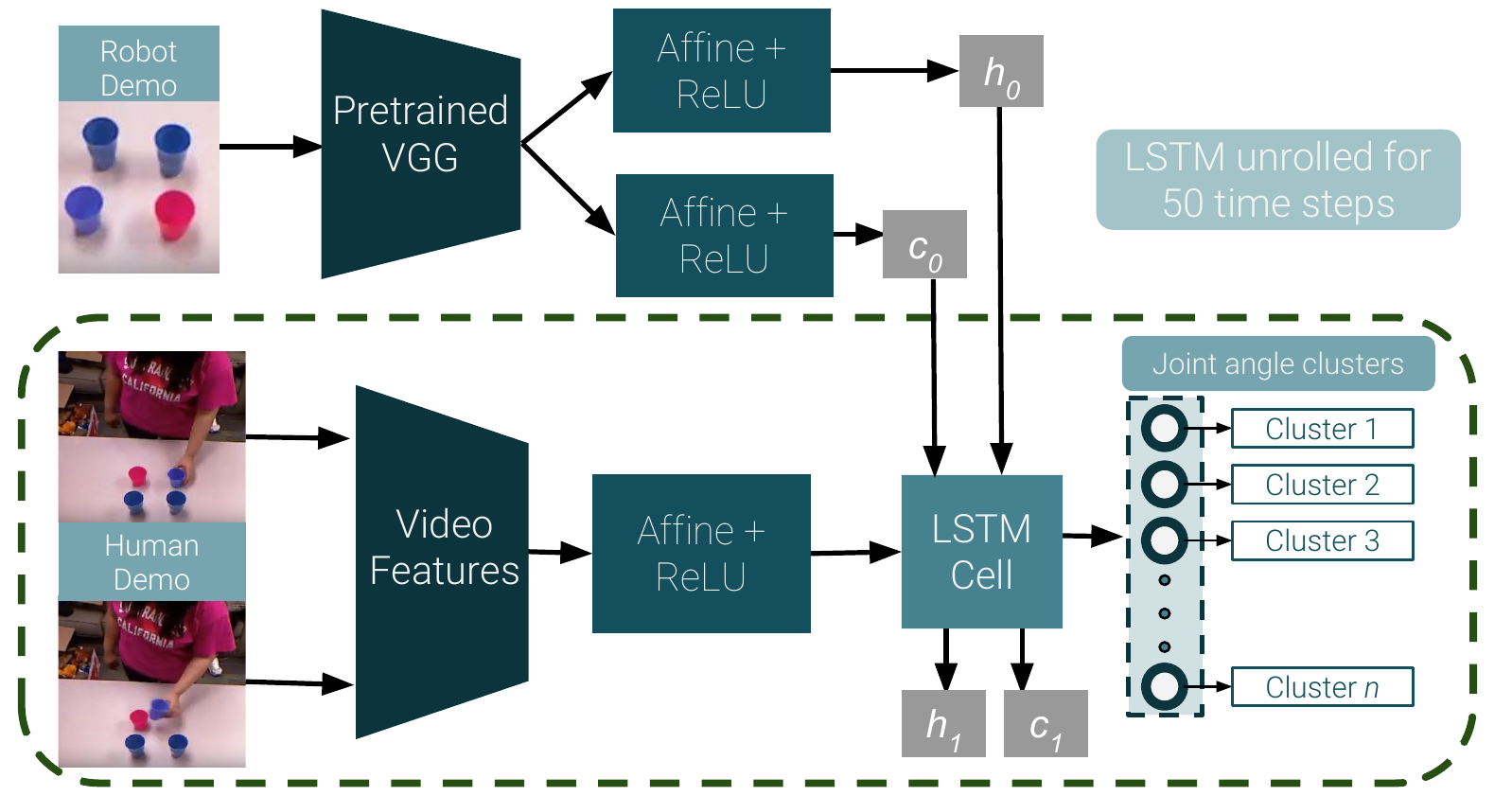}
\caption{Behavioural Cloning - Model Architecture.}
\label{fig:model}
\end{center}
\end{figure}

\noindent \textbf{Model Architecture:} Now that we have explained our input and output space, we explain our final model. Our model is visualized in Fig.~\ref{fig:model}. The goal of this model is to predict the joint angles $s_t$ given the original configuration of objects and a human video feature at every time-step. Hence this LSTM model runs across the length of the human video and gives us the predicted trajectory corresponding to the task in the human video. The LSTM's initial hidden state and cell state is set as the VGG feature of the robot demonstration image. The human demonstration video features are then used as input into the LSTM sequentially for 50 episodes. The loss for this network is the cross entropy loss between the predicted joint cluster-number and the ground truth joint cluster-number.





\noindent \textbf{Evaluation:} Next, we describe our evaluation metric for comparing predicted trajectories with the ground-truth trajectories for test videos. We use the Mean Squared Error (MSE) between the predicted trajectory and the true trajectory. We argue that though the mean square error might not be the best metric to evaluate the performance of the trajectory it turns out that using the MSE to evaluate the performance of a trajectory does help us evaluate how close a predicted trajectory is to one possible ground-truth. Individual classwise MSE errors are summarized in Table~\ref{tab:taskwisemse}. It can be seen that tasks that are multi-modal in nature, like placing in box, incur a larger loss.

\begin{table}[htb]
\centering
\caption{Task-wise MSE on held out test set}
\begin{tabular}{lllll} \toprule
Task & MSE &  \\\midrule
Pouring          &  0.111    &  \\ 
Stirring         &  0.1061     &   \\ 
Passing   & 0.1396      &   \\ 
Stacking         &   0.106    &   \\ 
Placing in a box & 0.1403       &     \\ 
Opening bottles  &   0.1245    &  \\ 
Pushing       &  0.1325      &   \\ 
Rotating objects &  0.1024     &    \\ 
Wiping with cloth           &  0.1435     &   \\ 
Typing on keyboard          &  0.1149     &   \\
\end{tabular}
\quad
\begin{tabular}{lllll} \toprule
Task & MSE &   \\\midrule
Close book          &   0.83487    &  \\ 
Single hand picking         & 0.1262      &  \\ 
Both hand picking   &  0.1172     &  \\ 
Poking         &  0.1179     &  \\ 
Two hand pull & 0.1066      &   \\ 
Two hand push  &   0.1049    &   \\ 
Toy car trajectories       &    0.1206   &  \\ 
Rolling cylinders & 0.1236      &   \\ 
Dropping           &   0.1211    &    \\ 
One hand pull           & 0.1506      &  \\
\end{tabular}
\label{tab:taskwisemse}
\end{table}

\textbf{Multimodality in the trajectory predicted:} A manipulation task can be solved by multiple different trajectories. There could be differences in the trajectory followed to reach the object, differences in location of grasping, or differences due to where the object was left after manipulation. Hence, given a state of the environment, there exist several possible trajectories that can achieve the task. To handle this we took two steps. The steps being classification instead of regression (which was discussed earlier in this section) and evaluation against multiple ground truths. In the following paragraph we discuss the second step.

If there are multiple possible trajectories and an approach selects one of them but the GT is the some other; then the approach gets penalized even though it was correct. To handle this, we collected multiple ground truth trajectories (while more is better we collected 2 in this paper) for all the tasks in the test set. The MSE was then calculated as the minimum of the MSE between the predicted trajectory and the each of the ground truth trajectories. Using this multiple GT trajectories, our MSE fell from 0.1296 to 0.1076. 

\begin{figure}[htbp]
\begin{center}
\includegraphics[width=4.2in]{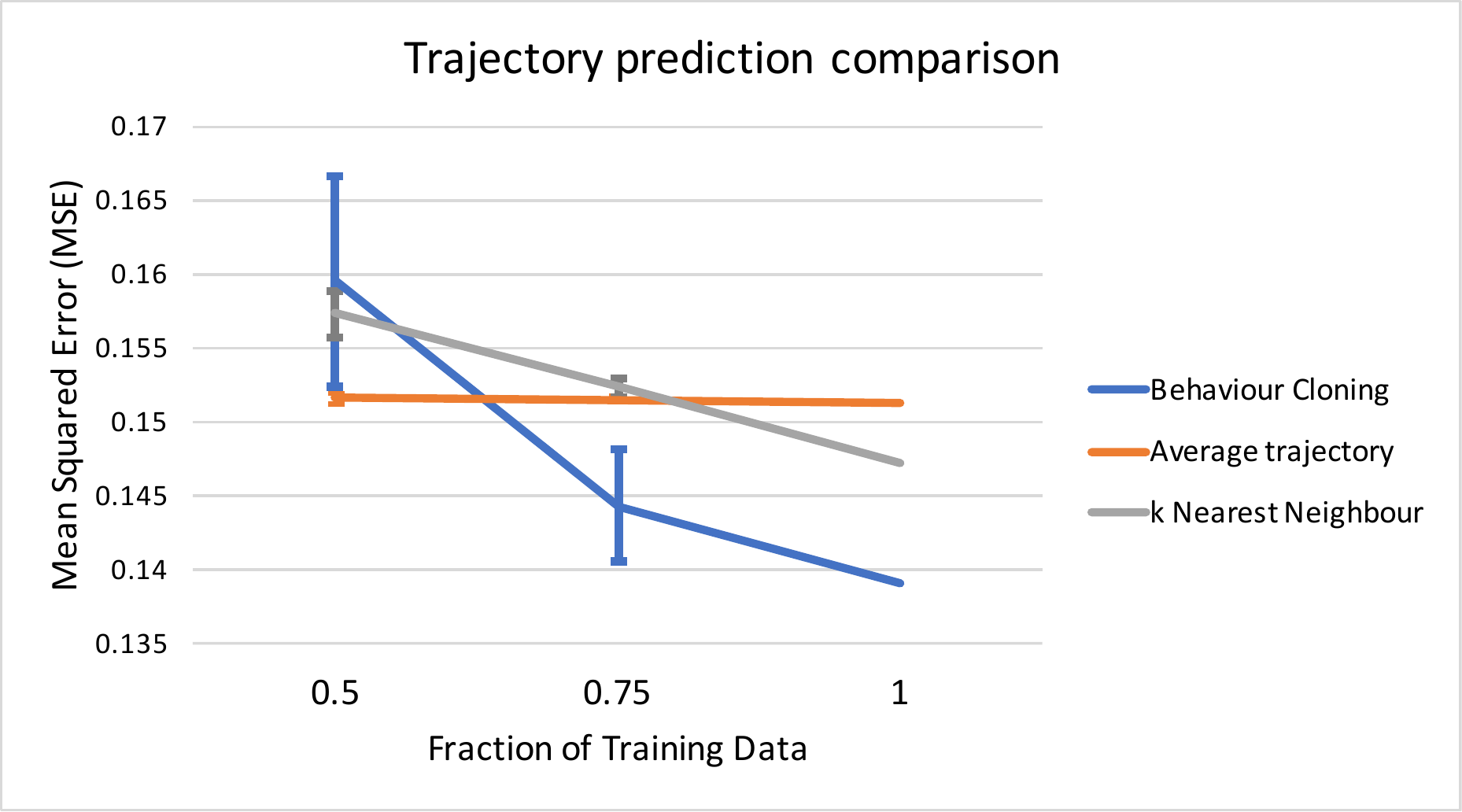}
\caption{Evaluation of behavior cloning, average trajectory and k-NN approach. The error bars show $\pm 1$ standard deviation when training on random splits with a fixed fraction of training data. }
\label{fig:comparision}
\end{center}
\end{figure}

\subsection{Comparing baselines}
We also implemented some other simple baselines to demonstrate how hard the dataset is. Specifically, we tried two baseline methods. 

\textbf{Average trajectory:} In this experiment, we used the average trajectory for each task and use it to compare with GT trajectory. 

\textbf{k-nearest neighbours:} As a second baseline we use Nearest-Neighbor baseline. Specifically, we compute the video features of the query human demonstration and then use it to retrieve the k-NN robot trajectories. The MSE loss was computed between the predicted trajectory against the multimodal ground truth trajectories. For our experiments we use k=11 as it yielded the best performance.

Fig.~\ref{fig:comparision} illustrates the variation of MSE with respect to the amount of data used. It can be seen that the behaviour cloning provides increasing performance with increasing amount of data used to train it. The method also does better than the two baselines while having a steeper improvement in performance with increase in data.


\section{Conclusion}
\label{sec:conclusion}
In this paper, we have presented one of the largest human and robot demonstration dataset to date. Our dataset consist of 8260 human-object interactions and 8260 robot trajectories for the same task on 20 diverse tasks. We demonstrate the use of our dataset for the task of visual imitation: mapping 3rd-person video features to robot trajectories.

\clearpage

{\footnotesize
\bibliography{references.bib}  
}
\end{document}